\documentclass[conference]{IEEEtran}
\IEEEoverridecommandlockouts
\usepackage{cite}
\usepackage{amsmath,amssymb,amsfonts}
\usepackage{algorithmic}
\usepackage{graphicx}
\usepackage{textcomp}
\usepackage{xcolor}
\usepackage[hidelinks]{hyperref}
\def\BibTeX{{\rm B\kern-.05em{\sc i\kern-.025em b}\kern-.08em
    T\kern-.1667em\lower.7ex\hbox{E}\kern-.125emX}}
\begin{document}

\title{Volatility-Aware Extreme Event Detection in High-Frequency Financial Markets\\
}

\author{
\IEEEauthorblockN{1\textsuperscript{st} Maorufa Zaman}
\IEEEauthorblockA{\textit{Department of Digital Business \& Innovation} \\
\textit{Tokyo International University}\\
Saitama, Japan \\
s22224551@al.tiu.ac.jp}
~\\
\and
\IEEEauthorblockN{2\textsuperscript{nd} Haris Md Sahed}
\IEEEauthorblockA{\textit{Department of Digital Business \& Innovation} \\
\textit{Tokyo International University}\\
Saitama, Japan \\
s22224542@al.tiu.ac.jp}
}
\maketitle

\begin{abstract}
Predicting extreme price movements in high-frequency financial markets is a challenging task due to non-stationarity, heavy-tailed return distributions, and severe class imbalance. In particular, rare but impactful events are often difficult to detect using conventional modeling approaches, which typically treat extreme movements as isolated observations. 
This study proposes a volatility-aware approach for extreme event detection using high-frequency Bitcoin limit order book (LOB) data. Motivated by empirical evidence of volatility clustering, the target formulation is extended to incorporate both large future returns and high-volatility regimes. This redefinition increases the proportion of informative samples and aligns the learning objective with the underlying market dynamics.

Using a tree-based model (XGBoost) with time-series cross-validation and imbalance-aware evaluation, the proposed method achieves a Precision-Recall AUC of approximately 0.40, significantly outperforming the baseline formulation with a PR-AUC of around 0.06. This represents more than a sixfold improvement in detecting rare events.

The results highlight that target design plays a critical role in financial machine learning, often exceeding the impact of model complexity. By incorporating volatility structure into the labeling process, the proposed approach provides a more effective and realistic framework for extreme event detection in high-frequency cryptocurrency markets.
\end{abstract}

\begin{IEEEkeywords}
Limit order book (LOB), High-frequency trading, Machine learning, XGBoost, Non-stationary data, Class imbalance, Time-series forecasting, Financial data analytics
\end{IEEEkeywords}
\section{Introduction}

Predicting short-term price movements from limit order book (LOB) data is a fundamental problem in financial machine learning, with direct applications in algorithmic trading, risk management, and market analysis. The LOB provides a real-time view of market microstructure by recording buy and sell orders across multiple price levels, motivating a broad class of machine learning approaches for capturing short-term price dynamics \cite{b1}.

Models trained on raw LOB features often exhibit limited predictive performance under realistic evaluation conditions. Three data-level properties drive this difficulty. First, LOB data are strongly non-stationary: statistical properties shift as market regimes evolve, causing models to generalize poorly to new conditions \cite{b2}. Second, microstructure noise from rapid order updates and cancellations obscures predictive signals. Third, the class distribution is severely imbalanced: in the Bitcoin LOB dataset used here, fewer than 2\% of observations qualify as extreme events under a standard quantile-based definition.

Much existing work responds to these challenges by advancing model architecture, from recurrent networks and convolutional models to hybrid forecasting systems \cite{b3}. However, architectures that perform well on benchmark datasets frequently fail under non-stationarity and severe class imbalance \cite{b4}, and complex models do not consistently outperform simpler alternatives across different assets and forecasting horizons \cite{b5}. Predictive performance is therefore shaped not only by model capacity, but also by how the prediction target is defined, how dataset properties are addressed, and whether the evaluation protocol reflects realistic temporal structure.

Related research has also investigated the application of machine learning and deep learning techniques to extreme-event forecasting across different domains. For example, a previous study by one of the authors evaluated machine learning and deep learning models, including Random Forest, XGBoost, and LSTM, for extreme climate event forecasting \cite{b6}. Building on this research experience in data-driven extreme-event modelling, the present study investigates a different but related problem: detecting extreme events in high-frequency financial markets through volatility-aware target formulation.

Robust out-of-sample evaluation is equally important. Static train-test splits can produce overly optimistic estimates by ignoring structural change and temporal dependency \cite{b7}, and rolling-window validation is essential when model parameters are time-varying \cite{b8}.

This paper addresses the formulation problem directly. Rather than proposing a new architecture, we take a data-centric perspective and frame the task as binary classification of rare extreme price movements in high-frequency Bitcoin LOB data. We show that a standard quantile-based event definition yields performance close to the random baseline, and that a volatility-aware target incorporating high-volatility regimes alongside large return magnitudes produces substantial improvements in PR-AUC.

Differencing-based stationarization, imbalance-aware learning, threshold calibration, and time-series cross-validation are treated as a unified pipeline rather than independent choices.

The main contributions of this study are as follows: (1) empirical demonstration that raw LOB features yield near-random predictive performance under realistic evaluation; (2) identification of target misspecification as an important bottleneck alongside non-stationarity and class imbalance; (3) a diagnostic-driven framework that adapts target formulation, preprocessing, and evaluation to dataset properties; and (4) observations across multiple temporal resolutions suggest that problem formulation, rather than model complexity, drives robust performance.
```
\section{Literature Review}
\subsection{Why Cryptocurrency Market Prediction Matters}

Predicting asset prices in electronic markets is a central problem in quantitative finance, with
direct implications for trading strategy, risk management, and market-making. The limit order
book (LOB) continuously records supply and demand across multiple price levels and has
emerged as a rich data source for modeling short-term price dynamics \cite{b1}. Unlike aggregate
price series, it encodes microstructure detail that is directly relevant to high-frequency
prediction tasks.
The growing participation of retail investors, institutional funds, and algorithmic systems has
amplified both the opportunities and the difficulties in cryptocurrency markets. Volatility creates
exploitable opportunities for well-calibrated models \cite{b2}, particularly when it comes to detecting
rare but impactful price movements. Cryptocurrency markets are inherently non-stationary and
nonlinear, driven by sentiment, regulatory events, macroeconomic shifts, and speculative
behavior \cite{b9}.
High-frequency LOB data introduces further complexity. Cross-level dependencies are both
spatial and temporal, and meaningful signals are buried within substantial microstructure noise
\cite{b10}. Decentralized market structure and variable liquidity only compound these difficulties. It is
against this backdrop that machine learning approaches have attracted sustained research
attention.
\subsection{From Statistical Baselines to Machine Learning}
ARIMA and GARCH-type models have long served as standard benchmarks for financial time-series forecasting \cite{11}. ARIMA captures linear autocorrelation under stationarity, while GARCH accounts for time-varying volatility. Both assumptions are reasonable in principle, but empirical studies consistently confirm that cryptocurrency markets routinely violate them \cite{b12}. Comparative studies suggest that classical models tend to be outperformed by nonlinear alternatives—such as support vector regression and shallow neural networks—when data exhibits complex, regime-dependent structure \cite{b5}. This pattern is characteristic of most cryptocurrency trading environments, where the distributional assumptions underlying classical methods are frequently violated. These limitations motivated the broader adoption of tree-based and kernel methods. Random Forest, XGBoost, and SVM/SVR gained traction because they impose fewer distributional assumptions and can accommodate large feature sets without extensive preprocessing. Multi-model comparison studies indicate that model rankings remain context-dependent, varying across assets, time horizons, and market regimes \cite{b4}. Research on hybrid LSTM-GARCH architectures suggests that, even when GARCH parameters are incorporated directly as features, simpler models can achieve comparable predictive accuracy under certain data conditions \cite{b5}. Related work has also explored hybrid GARCH-ANN frameworks for cryptocurrency volatility prediction under non-Gaussian market conditions [12]. Yu et al. \cite{b14} demonstrated that preprocessing and signal decomposition alone can meaningfully improve forecasting outcomes in complex financial environments. These findings reinforce the view that predictive performance is shaped not only by model selection, but also by how data is structured and presented to the model.

\subsection{Deep Learning Approaches and Their Limitations}
Recurrent architectures—including LSTMs, GRUs, and bidirectional variants—have been widely applied to cryptocurrency forecasting, with reported improvements in accuracy over both classical and shallow methods \cite{b3}. Extensions incorporating attention mechanisms and convolutional components address temporal and structural dependencies alike \cite{b3}. More recently, Transformer-based architectures have been explored in financial time-series forecasting, though their computational cost and data requirements limit their usefulness in high-frequency, low-sample settings \cite{b3}. For LOB-specific tasks, models such as DeepLOB have demonstrated strong mid-price prediction accuracy on high-frequency order book data \cite{b15}. Research on benchmark LOB datasets further suggests that deep learning can capture signals from order flow that simpler methods consistently fail to identify \cite{b16}. A GRU-based framework applied to Bitcoin price prediction—using price volatility, hash rate, and market sentiment as features—found that GRU tends to outperform LSTM when training data is limited \cite{b17}. This is a particularly relevant finding for high-frequency extreme event prediction, where labeled observations are sparse. These advances come with real practical limitations. Such models demand large training sets, extensive hyperparameter search, and careful feature engineering. Comparison studies indicate that rankings across model classes are inconsistent across assets and evaluation windows, and that ensemble approaches do not reliably outperform the best individual model \cite{b15}. There is also a tendency to evaluate architectures under controlled conditions that do not reflect the distributional shifts or label sparsity encountered in real deployment \cite{b17}. Reported gains can be difficult to replicate once the data distribution shifts or labeled events become rare. These observations suggest that increasing architectural sophistication alone may not consistently improve performance in imbalanced, high-frequency settings. Deep learning nonetheless remains valuable where large-scale, stable datasets are available and continues to represent an important direction in financial forecasting.
\subsection{Data-Level Challenges: The Underappreciated Half of the Problem}
Despite widespread acknowledgment, data-level challenges in LOB prediction are rarely addressed in a coordinated way. Non-stationarity, microstructure noise, and class imbalance are typically noted in preprocessing sections and then set aside in favor of architectural contributions \cite{b3}. How much of the observed performance variability across models is actually attributable to data handling—rather than model design—remains underexplored. Class imbalance is among the most consequential of these issues. Price-moving events are comparatively rare in cryptocurrency markets, while most observations correspond to periods of minimal activity. Resampling strategies such as SMOTE and hybrid under/oversampling have been applied in related financial tasks, including credit risk assessment and cryptocurrency prediction \cite{b18}. However, these interventions are typically aimed at improving evaluation metrics, rather than examining how imbalance interacts with model stability or generalization. Combining SMOTE with Adaboost-SVM ensembles and time weighting has been shown to improve minority-class recall in financial distress prediction \cite{b18}. This offers a useful methodological reference for jointly optimizing oversampling and ensemble strategies under imbalanced conditions, though such approaches remain largely isolated rather than embedded in a principled pipeline. Evaluation methodology presents a further, related concern. Accuracy continues to be reported as a primary metric in imbalanced settings, where a classifier that predicts only the majority class can still appear to perform well \cite{b4}. Precision-recall AUC and F1-score provide far more informative assessments for rare extreme event detection. Standard train-test splits also disregard the temporal ordering of financial data, producing optimistic estimates that are unlikely to hold under realistic deployment conditions. Tashman \cite{b7} showed that rolling-origin evaluation provides more reliable forecasting assessment than fixed splits. Inoue et al. \cite{b8} further demonstrated that rolling-window selection is essential under the time-varying parameter conditions commonly observed in financial markets. A particularly underexplored issue concerns target formulation. Most studies define extreme events using fixed return thresholds or simple directional labels, without reference to the volatility regime in which an event occurs \cite{b9}. Financial markets exhibit well-documented volatility clustering, where extreme movements tend to concentrate in persistent high-variance regimes rather than appearing as isolated occurrences. Assigning labels based solely on return magnitude—without accounting for surrounding market conditions—produces target formulations that may be misaligned with the true structure of market risk. This misspecification makes the learning problem harder, particularly in high-frequency settings where the class distribution is already severely skewed. Across imbalance handling, stationarization, evaluation, and target formulation, existing work applies techniques in isolation rather than as coordinated components of a pipeline. No established approach exists for diagnosing dataset-specific properties and selecting appropriate interventions accordingly. This gap limits the reproducibility and generalizability of reported findings.
\subsection{The Role of Sampling Frequency}

Research in this area spans a wide range of temporal resolutions, from daily price series \cite{b17} and hourly data \cite{b2} to tick-by-tick LOB streams \cite{b15}. Model performance can be sensitive to this choice. High-frequency data preserves microstructure information that coarser resolutions discard, and the most appropriate model class may differ substantially across settings \cite{b10}. Temporal resolution shapes the character of predictive signals available to a model, with microstructure-driven patterns potentially weakening under aggregation. Less attention has been paid to how sampling frequency interacts with other dataset properties. A one-minute and a five-minute series derived from the same underlying LOB are likely to differ not only in granularity but also in their non-stationarity profile, noise level, and class distribution. Most studies treat their chosen resolution as sufficient without examining what may be lost or gained by varying it. This reinforces the broader observation that data representation choices—of which temporal resolution is one example can be as consequential for predictive outcomes as model selection.

\subsection{Research Gaps}

Research in cryptocurrency LOB prediction has progressed from linear statistical models to ensemble methods and, more recently, to deep learning and hybrid approaches. Each generation has produced measurable improvements under controlled conditions, yet out-of-sample performance remains inconsistent—particularly in the presence of non-stationarity, noise, and class imbalance \cite{b15}. This persistent gap suggests that increasing representational capacity alone is not enough. Evidence from multi-model comparison studies indicates that simpler approaches frequently match more complex ones across multiple assets and evaluation windows \cite{b15}. Evaluation methodology and data representation may be just as consequential for performance as model depth, which motivates a formulation-first rather than architecture-first approach. Three interrelated limitations characterize the existing literature. First, data-level challenges—including non-stationarity, microstructure noise, and class imbalance—are widely acknowledged but typically addressed through ad hoc preprocessing. How these factors interact, and which interventions are most effective, remains largely unexamined \cite{b3}. Second, evaluation practices are frequently misaligned with the task. Accuracy-based reporting obscures classifier behavior under imbalance, and standard train-test splits violate the temporal ordering of financial data [6]. Rolling-window selection has been shown to be essential under time-varying parameter conditions \cite{b8}, yet its adoption remains inconsistent. Third, target formulation most commonly relies on fixed return thresholds that ignore volatility regime structure, producing labels that may not reflect the true dynamics of market risk \cite{b9}. This study addresses all three limitations through a diagnostic-driven pipeline. Rather than proposing a new architecture, the work treats dataset characterization as the foundational step in model design. Key properties—including non-stationarity, volatility clustering, and class imbalance—are diagnosed explicitly, and each finding informs a corresponding design decision: differencing-based stationarization, a volatility-aware target formulation incorporating high-volatility regime indicators alongside return-based thresholds, imbalance-aware learning, threshold calibration via precision-recall analysis, and time-series cross-validation. This integrated approach ensures that preprocessing, labeling, and evaluation choices are grounded in observed dataset properties rather than applied as generic defaults. Empirical results suggest that this approach can improve robustness and predictive performance relative to standard problem formulations. These findings do not diminish the importance of architectural innovation. Rather, they indicate that principled target formulation and rigorous evaluation are essential prerequisites for reliable performance in high-frequency, imbalanced settings.
\section{Data Description and Exploratory Analysis}

The dataset consists of high-frequency Bitcoin limit order book (LOB) data containing 17,113 observations and 156 features. The features include midpoint price, bid-ask spread, trading activity (buy and sell volumes), and multiple depth-level order book attributes capturing market microstructure.

The data spans a period from April 7, 2021 to April 19, 2021, covering approximately 12 days of trading activity. Observations are recorded at a near-uniform 1-minute frequency, with an average interval of 60.24 seconds and minor variations due to system-level timing offsets. No missing values or duplicate entries are present, indicating high data quality suitable for time-series modeling.

\subsection{Price and Return Dynamics}

\begin{figure}[!t]
\centering
\includegraphics[width=0.9\columnwidth]{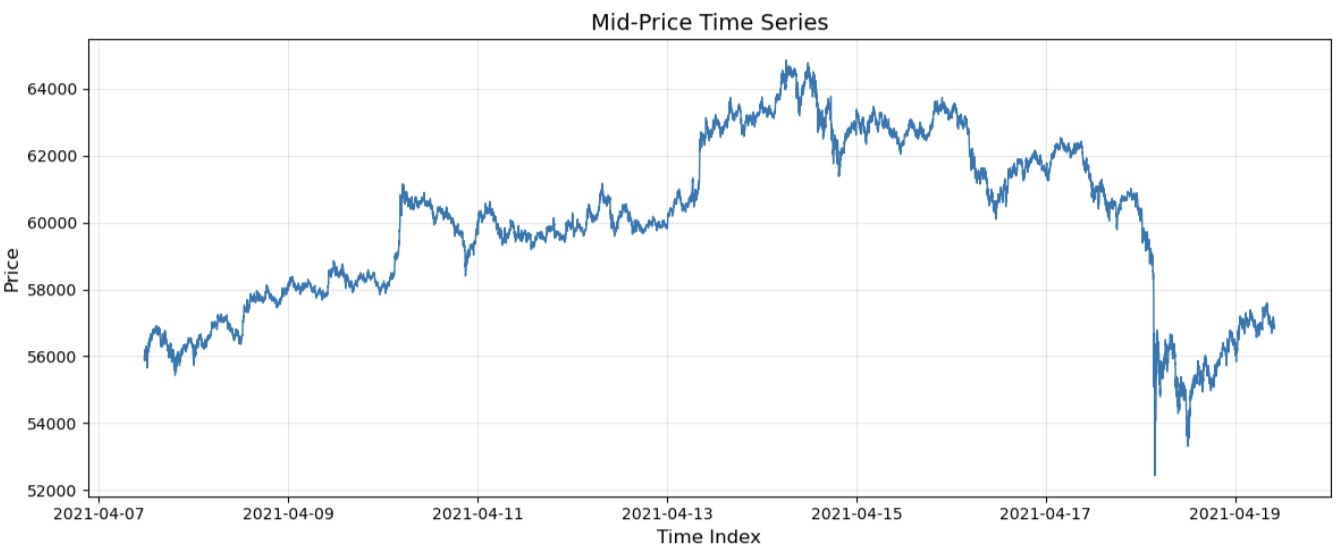}
\caption{Mid-price time series illustrating non-stationary behavior with trend shifts and abrupt movements.}
\label{fig:midprice}
\end{figure}

The mid-price series exhibits clear non-stationary behavior, as shown in Fig.~\ref{fig:midprice}, with multiple trend regimes, gradual drifts, and abrupt price movements. These characteristics indicate evolving market conditions and structural breaks.

\begin{figure}[!t]
\centering
\includegraphics[width=0.9\columnwidth]{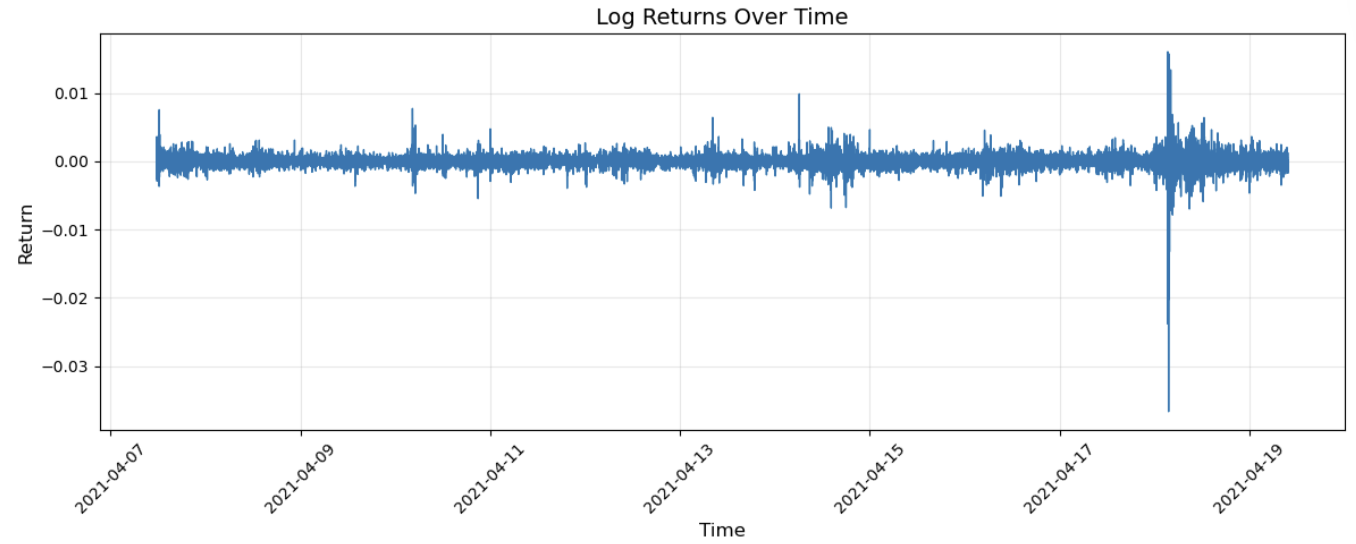}
\caption{Log returns of the mid-price showing frequent small fluctuations and occasional extreme spikes.}
\label{fig:returns}
\end{figure}

To analyze short-term dynamics, log returns are computed from the midpoint price. The return series, shown in Fig.~\ref{fig:returns}, has a near-zero mean ($1.00 \times 10^{-6}$) and a standard deviation of $0.00107$, but exhibits strong deviations from normality. Specifically, the distribution shows significant negative skewness ($-3.03$) and extremely high kurtosis (133.7), indicating a heavy-tailed distribution with frequent small fluctuations and occasional extreme movements.

These properties confirm that the data are non-Gaussian and dominated by rare but impactful events, which are critical for financial prediction tasks.

\subsection{Volatility Clustering}

Further analysis reveals strong volatility clustering, a well-known characteristic of financial time series. The rolling volatility (window size = 20) has a mean of $8.23 \times 10^{-4}$ and varies substantially over time, indicating alternating periods of low and high variance.

This behavior suggests temporal dependence and heteroskedasticity, violating the assumption of independent and identically distributed (i.i.d.) observations. Importantly, large return magnitudes tend to cluster together, forming high-volatility regimes.

\subsection{Class Imbalance and Target Distribution}
\begin{figure}[!t]
\centering
\includegraphics[width=0.9\columnwidth]{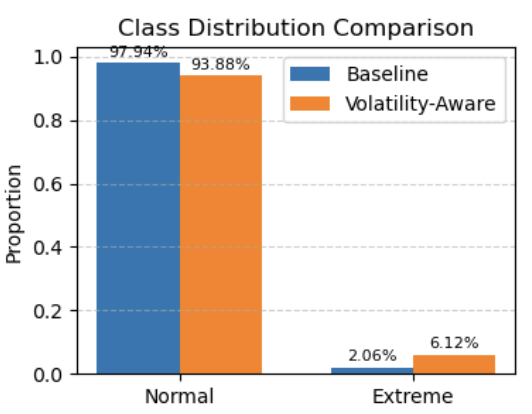}
\caption{Class distribution before and after target redefinition. The volatility-aware formulation increases the proportion of extreme events from approximately 2\% to 6\%, reducing class imbalance.}
\label{fig:imbalance}
\end{figure}
Extreme events are initially defined based on large future returns, resulting in a highly imbalanced dataset. The baseline distribution shows that approximately 97.95\% of observations correspond to normal market behavior, while only about 2.05\% represent extreme events.

To address this sparsity, a volatility-aware target is introduced, combining return-based extremes with high-volatility regimes. This increases the proportion of extreme events to approximately 6.10\%, as shown in Fig.~\ref{fig:imbalance}, providing a more informative and learnable target distribution. This modification improves the availability of positive samples, enabling the model to learn meaningful patterns associated with extreme events.
\section{Methodology}

\subsection{Problem Formulation}

The objective of this study is to predict short-term extreme price movements in high-frequency Bitcoin limit order book (LOB) data. Rather than modeling all price changes, which are often dominated by noise, we focus on identifying rare but significant market events.

Let $p_t$ denote the midpoint price at time $t$. The log return is defined as:
\begin{equation}
r_t = \log(p_t) - \log(p_{t-1})
\end{equation}

To capture short-term dynamics while reducing microstructure noise, a cumulative future return over a horizon of $H$ time steps is defined as:
\begin{equation}
R_t^{(H)} = \sum_{i=1}^{H} r_{t+i}
\end{equation}

In this study, we set $H = 3$ minutes to balance responsiveness and noise reduction.

\subsection{Baseline Extreme Event Definition}

A baseline definition of extreme events is constructed using a quantile-based threshold. Specifically, a symmetric threshold $\tau$ is defined as:
\begin{equation}
\tau = \text{Quantile}_{0.995}(|R_t^{(H)}|)
\end{equation}

An observation is labeled as extreme if:
\begin{equation}
|R_t^{(H)}| > \tau
\end{equation}

This formulation captures large price movements but results in a highly imbalanced dataset, with approximately 2\% of observations labeled as extreme. Moreover, it treats extreme events as isolated points, which does not reflect the temporal clustering observed in financial markets.

\subsection{Volatility-Aware Target Redefinition}

Empirical analysis of the dataset reveals strong volatility clustering, where periods of elevated variance persist over time. To incorporate this property, we extend the definition of extreme events by introducing a volatility-based regime indicator.

Rolling volatility is computed as:
\begin{equation}
\sigma_t = \text{Std}(r_{t-w+1}, \dots, r_t)
\end{equation}
where $w$ denotes the rolling window size.

A high-volatility threshold is defined as:
\begin{equation}
\sigma_{thr} = \text{Quantile}_{0.95}(\sigma)
\end{equation}

The final target variable is defined as:
\begin{equation}
y_t =
\begin{cases}
1, & \text{if } |R_t^{(H)}| > \tau \ \text{or} \ \sigma_t > \sigma_{thr} \\
0, & \text{otherwise}
\end{cases}
\end{equation}
This design reflects the fact that periods of elevated volatility often precede or accompany significant price movements, making them informative for extreme event detection.
This formulation captures both realized extreme movements and periods of elevated market risk. As a result, the proportion of positive samples increases from approximately 2\% to 6\%, reducing label sparsity while aligning the target with observed market dynamics.

\subsection{Feature Engineering}

To capture the temporal dynamics and microstructure characteristics of high-frequency markets, a diverse set of features is constructed using historical price and order flow information.

\subsubsection{Return-Based Features}
Lagged returns are included to capture short-term autocorrelation patterns:
\begin{equation}
r_{t-k}, \quad k \in \{1,2,3,5,10\}
\end{equation}
In addition, absolute returns and return differences are used to capture magnitude and acceleration effects.

\subsubsection{Volatility Features}
Rolling standard deviation of returns is computed over multiple windows to characterize time-varying volatility:
\begin{equation}
\sigma_t^{(w)} = \text{Std}(r_{t-w+1}, \dots, r_t), \quad w \in \{5,10,30,100\}
\end{equation}
A high-volatility regime indicator is further defined using a quantile threshold of long-window volatility.

\subsubsection{Momentum Features}
Momentum is captured using rolling sums of returns over different horizons:
\begin{equation}
\text{Momentum}_t^{(w)} = \sum_{i=0}^{w-1} r_{t-i}, \quad w \in \{3,5,7,10,30\}
\end{equation}
Short-term directional momentum is further encoded using lagged return signs and rolling ratios of positive returns.

\subsubsection{Order Flow and Imbalance Features}
Market pressure is represented using order flow imbalance:
\begin{equation}
\text{Imbalance}_t = \frac{\text{Buys}_t - \text{Sells}_t}{\text{Buys}_t + \text{Sells}_t}
\end{equation}
Additional features include imbalance changes and order pressure measures derived from buy and sell volumes.

\subsubsection{Spread-Based Features}
Bid-ask spread dynamics are incorporated through first differences and rolling volatility of the spread, capturing liquidity conditions and market frictions.

\subsubsection{Directional and Behavioral Indicators}
Rolling ratios of positive returns (e.g., proportion of upward movements) are used to capture short-term directional bias in price movements.

All features are constructed using only information available up to time $t$, ensuring that no future information is used and preventing look-ahead bias. These features are designed to capture volatility clustering and short-term dynamics associated with extreme events.
\subsection{Modeling Approach}

The extreme event detection task is formulated as a binary classification problem. We employ Extreme Gradient Boosting (XGBoost) due to its ability to capture non-linear relationships and effectively model tabular financial data under class imbalance.

To preserve temporal dependencies, model evaluation is performed using time-series cross-validation (TimeSeriesSplit), where training is conducted on past data and evaluated on future observations.

Class imbalance is addressed using the \texttt{scale\_pos\_weight} parameter, computed as the ratio of negative to positive samples within each training fold.

\subsection{Threshold Calibration}

Due to the rarity of extreme events, a fixed classification threshold (e.g., 0.5) leads to poor recall. To address this, threshold calibration is performed using the precision-recall curve.

The decision threshold is selected within each cross-validation fold using the evaluation split. Specifically, the threshold is chosen to maximize recall subject to a minimum precision constraint:
\begin{equation}
\max \ \text{Recall} \quad \text{s.t.} \quad \text{Precision} \geq \alpha
\end{equation}
where $\alpha = 0.15$ in this study. This reflects practical scenarios where missing extreme events is more costly than false positives.

To prevent degenerate solutions, a minimum threshold constraint is enforced:
\begin{equation}
t \geq t_{\min}
\end{equation}
where $t_{\min} = 0.01$. This constraint prevents unrealistically low thresholds that would trivially maximize recall at the expense of precision.

Threshold selection is performed independently for each cross-validation fold to account for temporal variation in the data.
\subsection{Evaluation Protocol}

Model performance is evaluated using time-series cross-validation to avoid look-ahead bias. The following metrics are reported:

\begin{itemize}
    \item Precision-Recall AUC (PR-AUC)
    \item F1-score
    \item ROC-AUC
\end{itemize}

PR-AUC is emphasized as the primary metric due to the imbalanced nature of the problem, as it better reflects the model's ability to identify rare extreme events.

\section{Results and Discussion}

\subsection{Experimental Setup}

The proposed approach is evaluated using time-series cross-validation, ensuring that models are trained on past data and tested on future observations. Performance is assessed using Precision-Recall AUC (PR-AUC), along with ROC-AUC, F1-score, precision, and recall.

Two target formulations are considered:

\begin{itemize}
    \item \textbf{Baseline Target:} Extreme events defined solely by large future returns (approximately 2\% positive class).
    \item \textbf{Volatility-Aware Target:} Extreme events defined using both large returns and high-volatility regimes (approximately 6\% positive class).
\end{itemize}

\subsection{Baseline Performance}

Under the baseline target formulation, model performance is
limited due to severe class imbalance and sparse labeling.
The PR-AUC is approximately 0.08, only slightly higher than
the random baseline ($\sim$0.06). In addition, recall remains
very low, indicating that the model fails to capture a
significant portion of extreme events.

These results suggest that defining extreme events solely
based on large returns may be insufficient for reliable
detection in highly imbalanced financial data.

\subsection{Performance with Volatility-Aware Target}

The proposed volatility-aware labeling leads to substantial
improvements in detection performance. By incorporating
high-volatility regimes, the positive class proportion
increases from approximately 2\% to 6\%, reducing label
sparsity and providing a more informative learning signal.

The model achieves the following average performance:

\begin{itemize}
    \item \textbf{PR-AUC:} $\sim$0.40
    \item \textbf{ROC-AUC:} $\sim$0.87
    \item \textbf{Precision:} $\sim$0.28
    \item \textbf{Recall:} $\sim$0.49
    \item \textbf{F1-score:} $\sim$0.31
\end{itemize}

Compared with the random baseline (PR-AUC $\sim$0.06),
the volatility-aware formulation achieves approximately
a 6.7-fold higher PR-AUC. It also achieves a fivefold
improvement over the baseline model (PR-AUC $\sim$0.08),
demonstrating the importance of target formulation in
addressing class imbalance and improving detection
performance.

Importantly, this improvement is achieved without changing
the model architecture, highlighting the influence of
target formulation on model performance.
\subsection{Impact of Threshold Calibration}

Threshold calibration further enhances detection performance. Instead of using a fixed threshold, the calibrated approach adjusts the decision boundary based on the precision-recall trade-off within each fold.

This allows the model to detect approximately 50\% of extreme events while maintaining reasonable precision. These results confirm that threshold selection plays a critical role in balancing false positives and missed events in imbalanced financial prediction tasks.
\subsection{Discussion}

The results demonstrate that the primary improvement arises from target reformulation rather than model complexity. Incorporating volatility information enables the model to capture regime-based behavior, which is a fundamental characteristic of financial markets.

These findings indicate that:

\begin{itemize}
    \item Extreme events are not isolated occurrences but part of broader volatility regimes.
    \item Proper label design significantly enhances model learnability.
    \item Tree-based models can effectively capture non-linear relationships when the target is well-defined.
\end{itemize}

Despite strong performance in extreme event detection, predicting the direction of such events remains challenging. Preliminary experiments on directional classification yielded results only marginally above the random baseline, indicating that short-term price direction is inherently more stochastic than volatility.

\section{Conclusion and Future Work}

This study presents a volatility-aware framework for detecting extreme price movements in high-frequency Bitcoin limit order book (LOB) data. By analyzing the statistical properties of the dataset, including non-stationarity, heavy-tailed return distributions, and volatility clustering, we identify key challenges that limit the effectiveness of conventional extreme event definitions.

To address these limitations, a volatility-aware target formulation is proposed, which integrates both large return magnitudes and high-volatility regimes. This approach increases the proportion of informative samples from approximately 2\% to 6\%, reducing label sparsity and improving model learnability. Combined with time-series cross-validation and threshold calibration based on precision-recall trade-offs, the proposed framework achieves substantial improvement in detecting rare but impactful market events.

Experimental results demonstrate that the proposed method significantly outperforms the baseline formulation, achieving a PR-AUC of approximately 0.40 compared to a near-random baseline of around 0.06. This highlights the importance of target design in financial machine learning, where the formulation of the problem can have a greater impact than the choice of model.

Despite these improvements, several limitations remain. The current study focuses on event detection and does not explicitly model the direction of price movements, which remains a challenging task in high-frequency financial data. Additionally, the dataset is limited in temporal coverage, which may restrict the generalizability of the findings.

Future work can extend this framework by incorporating larger and more diverse datasets, as well as higher-frequency order book information to capture finer-grained market dynamics. More advanced temporal models, such as deep learning architectures, may further improve the modeling of complex dependencies. Additionally, extending the framework to a multi-stage pipeline for both event detection and directional prediction remains a promising direction for further investigation.

Overall, the findings suggest that incorporating volatility structure into target design provides a more effective and realistic approach to extreme event detection in financial markets.

\end{document}